\documentclass[10pt,twocolumn,letterpaper]{article}

\usepackage{cvpr}              

\usepackage[dvipsnames]{xcolor}

\usepackage[normalem]{ulem} 

\definecolor{cvprblue}{rgb}{0.21,0.49,0.74}
\usepackage[pagebackref,breaklinks,colorlinks,citecolor=cvprblue]{hyperref}
\usepackage{multirow}
\usepackage{adjustbox}
\usepackage[accsupp]{axessibility} 

\newcommand{\Tref}[1]{Table~\ref{#1}}

\newcommand{\Fref}[1]{Figure~\ref{#1}}
\newcommand{\Sref}[1]{Sec.~\ref{#1}}

\def\paperID{2039} 
\def\confName{CVPR}
\def\confYear{2024}

\title{Language-guided Image Reflection Separation\vspace{-10pt}}

\def\instvspace{\vspace{-3pt}}

\author{Haofeng Zhong$^{1,2,3\,\#}$~~
Yuchen Hong$^{1,2\,\#}$~~
Shuchen Weng$^{1,2}$~~
Jinxiu Liang$^{1,2}$~~
Boxin Shi$^{1,2,3\,*}$\\
{\small $^1$ National Key Laboratory for Multimedia Information Processing, School of Computer Science, Peking University}\instvspace\\
{\small $^2$ National Engineering Research Center of Visual Technology, School of Computer Science, Peking University}\instvspace\\
{\small $^3$ AI Innovation Center, School of Computer Science, Peking University}\instvspace\\
{\small \texttt{\{hfzhong, shuchenweng, cssherryliang, shiboxin\}@pku.edu.cn, yuchenhong.cn@gmail.com}}
\vspace{-10pt}
}

\begin{document}
\maketitle

{
    \renewcommand*{\thefootnote}{$^{\#}$}
    \footnotetext{Equal contributions.~$^{*}$Corresponding author.}
}

\begin{abstract}
This paper studies the problem of language-guided reflection separation, which aims at addressing the ill-posed reflection separation problem by introducing language descriptions to provide layer content.
We propose a unified framework to solve this problem, which leverages the cross-attention mechanism with contrastive learning strategies to construct the correspondence between language descriptions and image layers.
A gated network design and a randomized training strategy are employed to tackle the recognizable layer ambiguity.
The effectiveness of the proposed method is validated by the significant performance advantage over existing reflection separation methods on both quantitative and qualitative comparisons.

\end{abstract}    
\section{Introduction}
\label{sec:intro}

When photographing through transparent materials like glass windows or showcases, the presence of reflections can significantly degrade the image quality of captured images and disrupt downstream computer vision tasks like face recognition~\cite{wan2021face} or depth estimation~\cite{chang2020joint}.
As an attractive topic in computational photography, reflection separation aims at decomposing the contaminated mixture image (denoted as $\mathbf{M}$) into two components that correspond to scenes located at different sides of the glass, \ie, the reflection layer (denoted as $\mathbf{R}$) and the transmission layer (denoted as $\mathbf{T}$).
Since reflection separation is a severely ill-posed problem, it is imperative to exploit effective priors or auxiliary information for distinguishing the two components.

\begin{figure}[t]
    \centering
    \includegraphics[width=\linewidth]{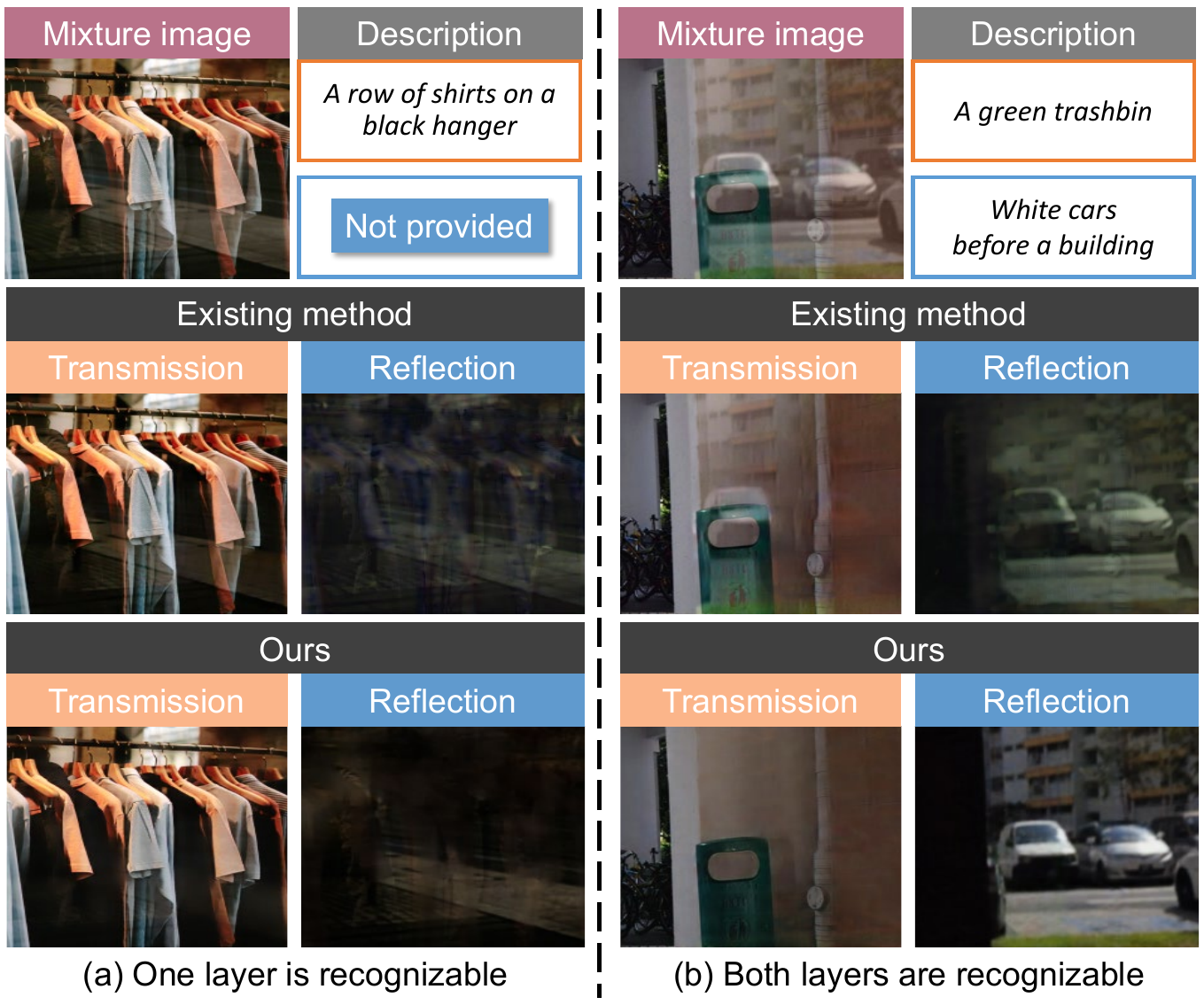}
    \vspace{-0.7cm}
    \caption{The recognizable layer ambiguity problem causes uncertain quantities of input language descriptions for language-guided image reflection separation.
    Given language descriptions of either (a) one layer or (b) both two layers, the proposed method achieves robust reflection separation compared with an existing reflection separation method~\cite{dong2021location}.}
    \label{fig:teaser}
\end{figure}

The primary challenge of solving the reflection separation problem lies in the exploration of distinct clues for distinguishing transmission and reflection layers.
Multi-image methods handle this problem by introducing additional constraints.
Some of them acquire a series of mixture images in different viewpoints~\cite{li2013exploiting,simon2015reflection,liu2020learning} to harness the distinct motions of the two layers, while others adopt specialized capturing setups to obtain complementary scene information~\cite{lei2020polarized,lyu2019reflection,lei2021robust,hong2021panoramic}.
However, the specialized data capture requirements limit the application scope of these methods, especially for images downloaded from the Internet.
Single-image methods attempt to tackle reflection separation by utilizing handcrafted priors derived from natural image statistics~\cite{levin2007user,li2014single,shih2015reflection} or leveraging the modeling capacity of neural networks to learn content priors about reflections from a large scale of training data~\cite{li2020single,dong2021location,hu2021ytmt}.
However, they are prone to fail due to the lack of auxiliary content information about transmission or reflection scenes for solving such a highly ill-posed problem.
Recently, language descriptions have shown their effectiveness in providing content information for various vision tasks such as image editing~\cite{sun2022unicorn,weng2023CIR,weng2020misc,tang2023luminaire}, semantic segmentation~\cite{yang2022lavt,wang2022cris}, and image colorization~\cite{weng2022lcode,chang2022coder,chang2023coins,chang2023lcad}, which inspires us to think about: \textit{Can we leverage the auxiliary content information brought by language descriptions to facilitate the reflection separation problem?}


Since language can effectively convey humans' prior knowledge about the real world~\cite{deng2023nerdi} and provide auxiliary information of image semantics~\cite{yang2022lavt}, introducing language descriptions to guide the separation of reflection and transmission layers from mixture images merits exploration.
However, leveraging language descriptions for reflection separation is non-trivial in three aspects:
\textbf{1) Language-image modality inconsistency}. Language and images belong to different modalities, thus it is challenging to establish a cross-modality correspondence between the scene content information provided in language descriptions and the complex blended content present in mixture images.
\textbf{2) Recognizable layer ambiguity}. Since the image content and brightness of reflection and transmission layers are different, the recognizable extents of them in mixture images are also uncertain.
Specifically, as shown in~\Fref{fig:teaser}, it is possible that only one layer's content is recognizable (clothes in \Fref{fig:teaser}(a)), or both two layers exhibit recognizable content (the trashbin and cars in \Fref{fig:teaser}(b)), which leads to the difficulty of using uncertain quantities of language descriptions for separating mixture images in practice.
\textbf{3) Language annotation deficiency}. All existing datasets for the reflection separation task only contain image data but no correlated language description is provided, raising the challenge for network training and evaluation.

In this paper, we introduce the concept of \textit{language-guided image reflection separation} for the \textit{first} time, which leverages flexible natural language to specify the content of one or two layers within a mixture image, relieving the ill-posedness of the reflection separation problem and maintaining a wide applicability for both live captured or online downloaded mixture images.
We propose an end-to-end framework that employs adaptive global interaction modules to explore holistic language-image content coherence and utilizes specifically designed loss functions to constrain the correspondence between language descriptions and recovered image layers.
A language gate mechanism and a randomized training strategy are designed to deal with the recognizable layer ambiguity problem.
To address the language annotation deficiency, we synthesize the training dataset from paired image-language datasets~\cite{young2014flickr30k,chen2015cococaption} and expand prevailing real reflection separation datasets~\cite{wan2017sir2,zhang2018single,li2020single} by manually adding language descriptions.
Besides, we further construct a new dataset for visual quality evaluation by collecting mixture images from the Internet and captioning them with language descriptions for recognizable layers.
Our contributions are summarized as follows:
\begin{itemize}
    \item We present the first work that introduces language descriptions to guide the reflection separation task.
    \item We propose adaptive global interaction modules and language-image loss functions to tackle modality inconsistency.
    \item We design a language gate mechanism and a randomized training strategy to handle recognizable layer ambiguity.
    \item We build a dataset with language descriptions to facilitate language-guided image reflection separation.
\end{itemize}
\section{Related work}

\textbf{Single-image reflection separation} methods try to distinguish reflection and transmission layers using a single mixture image, which mainly relies on the assumption that the two layers have different distributions, \ie, reflection layers are more likely to be blurry and appear with lower intensity compared with transmission layers.
Conventional methods adopt handcrafted priors derived from natural image statistics in their optimization process, \eg, the gradient sparsity~\cite{levin2007user}, relative smoothness~\cite{li2014single}, ghosting cues~\cite{shih2015reflection}, content prior~\cite{wan2018region}, and penalty on the gradient of restored transmission layers~\cite{yang2019fast}.

Due to the tremendous progress in the field of deep learning, a series of single-image reflection separation methods concentrate on the improvement of learning strategies or network design, \eg, predicting edges and images with a two-stage~\cite{fan2017generic} or concurrent framework~\cite{wan2018crrn,wan2019corrn}, training with the perceptual loss~\cite{zhang2018single}, employing generative adversarial network~\cite{goodfellow2014generative} based models~\cite{wei2019single,ma2019learning}, adopting iterative refinement strategies~\cite{yang2018seeing,li2020single,dong2021location,zhang2022content}, and leveraging the complementary two-stream architecture~\cite{hu2021ytmt}.
Meanwhile, research on data synthesis and image models is also ongoing to satisfy the data-driven needs of learning-based methods.
Ma~\etal~\cite{ma2019learning} utilize generative adversarial networks for data generation while Wen~\etal~\cite{wen2019single} synthesize mixture images with learned non-linear blending masks.
Hu~\etal~\cite{hu2023single} introduce a learnable residue term in the mixture image formation model to mitigate the non-linearity caused by the complex camera pipeline.
Zheng~\etal consider physical factors such as reflective amplitude coefficient maps~\cite{zheng2020what} and the absorption effect~\cite{zheng2021absorb} in the image formation process of mixture images.
To facilitate network training and evaluation, researchers~\cite{wan2017sir2,zhang2018single,li2020single} also collect real data by using portable glass.
Moreover, as a special form of images, panoramic images are introduced to relieve the content ambiguity in mixture images~\cite{hong2021panoramic,hong2023par2net,han2022zero,park2024fully}.
We refer readers to~\cite{wan2022sir2+} for a comprehensive and up-to-date survey on single-image reflection separation.


\begin{figure*}[t]
    \centering
    \includegraphics[width=\linewidth]{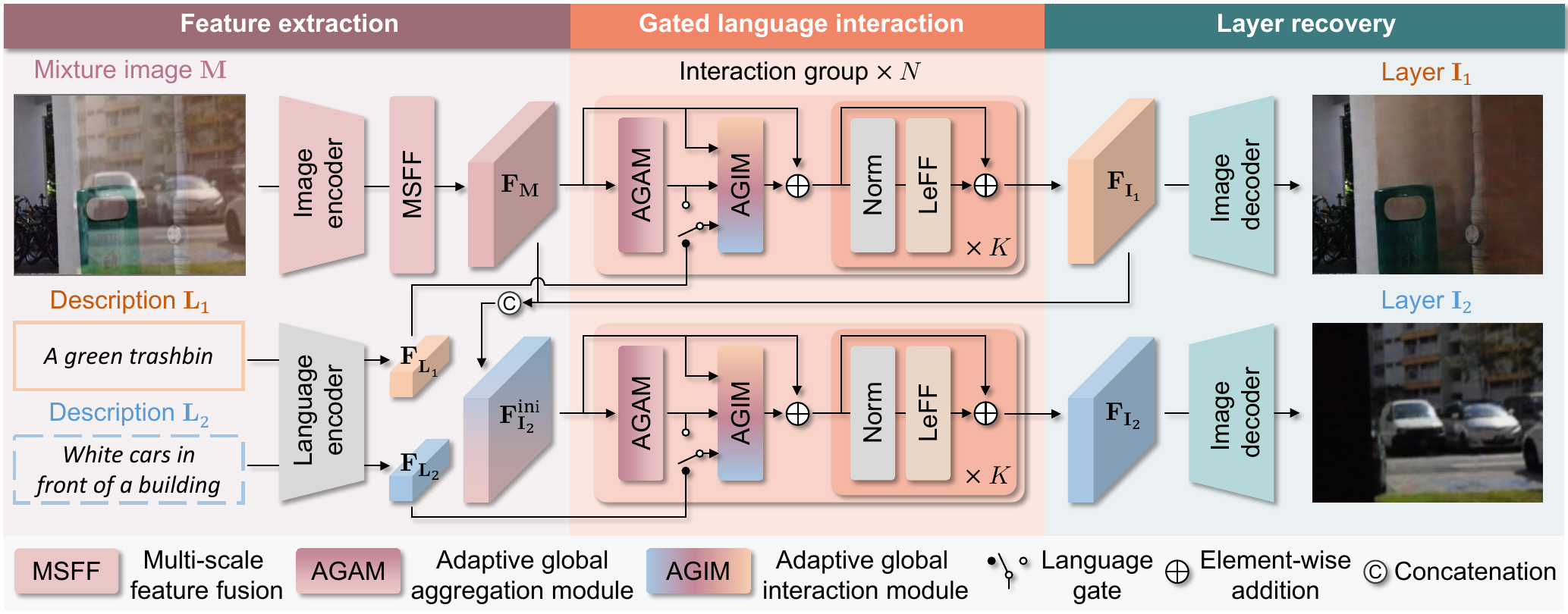}
    \vspace{-0.7cm}
    \caption{
    The pipeline of the proposed language-guided image reflection separation framework, which extracts features from mixture images and available language descriptions (the description $L_2$ with dashed lines is possible to be set to null due to the recognizable layer ambiguity) by image and language encoders (in \Sref{subsec:extract}), aggregates global visual information by adaptive global aggregation modules (AGAM) and conducts progressive interactions to exploit distinctive image features with gated language guidance by adaptive global interaction modules (AGIM) (in \Sref{subsec:interact}), and recovers image layers by image decoders (in \Sref{subsec:recover}).}
    \label{fig:pipe}
\end{figure*}

\noindent\textbf{Multi-image reflection separation} methods usually leverage the auxiliary information introduced by additional images and achieve more robust performance than single-image methods.
Polarization-based methods~\citep{nayar1997separation,schechner2000separation,diamant2008overcoming,kong2012physically,lyu2019reflection,lei2020polarized,lyu2022physics} distinguish reflection and transmission layers by using images captured with different angles of polarizers or special polarization cameras.
Flash-based methods~\citep{chang2020siamese,lei2021robust,lei2023robust,hong2020near,hong2022nir3net} adopt active light sources to illuminate transmission scenes for obtaining reflection-free guidance.
Motion-based methods~\citep{li2013exploiting,simon2015reflection,liu2020learning,liu2021learning} utilize multiple images captured from different viewpoints to harness distinct motions of reflection and transmission layers.
However, special data capture requirements significantly limit the application scope of these methods, especially for mobile devices or images downloaded from the Internet.

\section{Proposed method}
The pipeline of the proposed framework is illustrated in~\Fref{fig:pipe}.
In this section, we start by introducing the feature extraction stage (in \Sref{subsec:extract}) that obtains multi-scale image features and global language features, the gated language interaction stage (in \Sref{subsec:interact}) that conducts progressive image-language global interactions to exploit distinctive image features and prevents unavailable language interactions with switchable gates, and the layer recovery stage (in \Sref{subsec:recover}) that reprojects features into the image space with a light-weight image decoder.
Then we explain loss functions (in \Sref{subsec:loss}) employed for network optimization, especially a contrastive correspondence loss and a layer correspondence loss that constrains the network to construct correspondences between the language description and the corresponding layer under the disturbance of the other layer in a superimposed mixture image.
Finally, we present our training strategy (in \Sref{subsec:train}) which enables the network to be applicable for varying quantities of input language descriptions and jointly tackles the recognizable layer ambiguity problem with the gated network design.

\subsection{Image and language feature extraction}\label{subsec:extract}

The inputs of the proposed method consist of a mixture image $\mathbf{M}$ with two language descriptions $\{\mathbf{L}_i|i=1,2\}$ which corresponds to the two image layers.
We specify that layer $\mathbf{I}_i$ corresponds to the description $\mathbf{L}_i$.
However, due to the recognizable layer ambiguity that in certain cases only one layer of the mixture image is recognizable (usually the transmission layer), for such cases, we set $\mathbf{L}_1$ to be the available language description (for the recognizable transmission layer) and $\mathbf{L}_2$ to null (for the unrecognizable reflection layer) to ensure a unified input setting.
Then given the input image and language descriptions, the feature extraction stage aims at obtaining the image feature $\mathbf{F}_\mathbf{M}$ with the image encoder and the multi-scale feature fusion process and extracting the global language feature $\mathbf{F}_{\mathbf{L}_i}$ for each description $\mathbf{L}_i$ via the language encoder for the subsequent interaction procedure.



\noindent\textbf{Image encoder.}
Given a mixture image $\mathbf{M}\in\mathbb{R}^{H\times{W}\times{3}}$, we employ a commonly-used vision backbone ResNet-50~\cite{he2016deep} as our image encoder, whose last three layers (\ie, an average pooling layer, a fully connected layer, and a softmax layer) are removed to fit our task.
We utilize image features from the first five blocks of the image encoder to form a multi-scale feature pyramid $\{\mathbf{F}_{\mathbf{M}_i}\}_{i=1}^5$, where $\mathbf{F}_{\mathbf{M}_i}\!\in\!\mathbb{R}^{{h_i}\times w_i\times C_i}$, $h_i\!=\!H/{2^i}$ and $w_i\!=\!W/{2^i}$, $H$ and $W$ is the height and width of the mixture image, respectively, and $C_i$ is the dimension of the $i$-th extracted feature.

\noindent\textbf{Multi-scale feature fusion.}
Obtaining the extracted feature pyramid, we first transform it into a hypercolumn feature $\mathbf{F}^{\rm hyp}_{\mathbf{M}}\in\mathbb{R}^{h\times w\times C^{\rm hyp}}$~\cite{hariharan2015hypercolumns} (where $h\!=\!H/2$, $w\!=\!W/2$, and $C^{\rm hyp}\!=\!\sum_{i=1}^{5}C_i$), which has been proved to be effective in fusing multi-scale contextual information for reflection separation~\cite{zhang2018single,wei2019single}.
Considering the computational cost, we condense and refine the hypercolumn feature by a $1\times1$ convolutional layer with a GELU activation~\cite{hendrycks2016gelu} followed by a locally-enhanced feed-forward (LeFF) block~\cite{wang2022uformer}. 
The final fused feature of the mixture image is denoted as $\mathbf{F_{M}}\in\mathbb{R}^{h\times w\times C}$, which serves as the basis for the subsequent interaction and separation process. 

\noindent\textbf{Language encoder.}
Motivated by the rapid development of pre-trained large-scale vision-language models, we employ the language encoder from CLIP~\cite{radford2021CLIP}, which adopts a Transformer architecture~\cite{vaswani2017attention} to extract language features and obtains a global contextual feature in the multi-modal embedding space by using layer normalization and linear projection layers.
Given a language description $\mathbf{L}_i\in\mathbb{R}^L$, we obtain its corresponding global feature $\mathbf{F}_{\mathbf{L}_i}\in\mathbb{R}^C$ by leveraging the modeling capacity of the language encoder to encode the description, thus extracting the holistic contextual information of the corresponding image layer.
Here $L$ denotes the length of the language description and $C$ is the feature dimension as the image feature.




\begin{figure}[t]
    \centering
    \includegraphics[width=\linewidth]{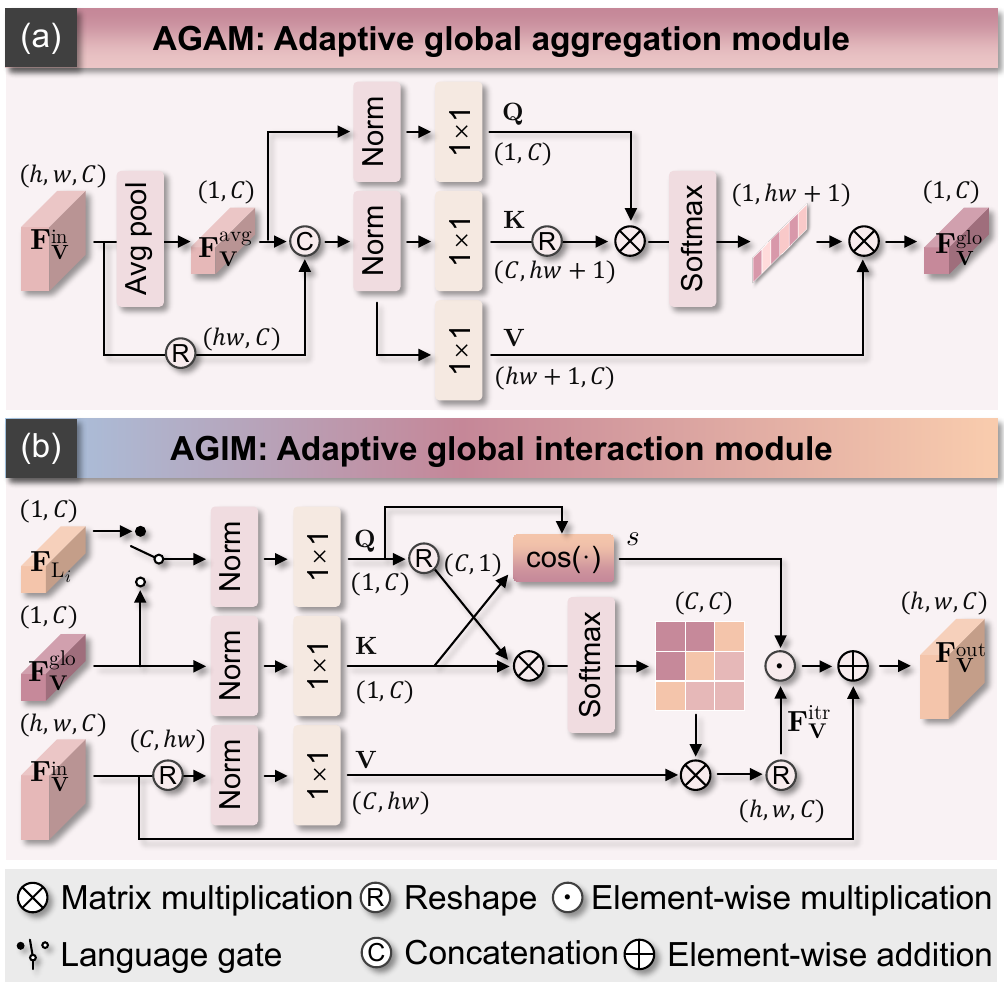}
    \vspace{-0.7cm}
    \caption{The architecture of the (a) adaptive global aggregation module (AGAM) and (b) adaptive global interaction module (AGIM), which aggregates global contextual information of visual features and achieves feature channel rearrangement with gated language guidance, respectively.}
    \label{fig:interaction}
\end{figure}

\subsection{Gated language interaction}\label{subsec:interact}
The gated language interaction stage aims at leveraging the contextual information from available language descriptions to guide the separation of the corresponding layer feature $\mathbf{F}_{\mathbf{I}_i}$, which is composed of $2N$ cascaded interaction groups to separate image layer features successively.
As shown in \Fref{fig:pipe}, each group consists of an adaptive global aggregation module (AGAM) to gather global information of input visual features, a language gate to prevent detrimental guidance from unavailable descriptions, an adaptive global interaction module (AGIM) to conduct interactions using global features for exploring holistic image-language content correspondence, and $K$ normalization layers with LeFF blocks~\cite{wang2022uformer} for feature refinement.
The former $N$ interaction groups are utilized for separating $\mathbf{F}_{\mathbf{I}_1}$ from $\mathbf{F}_\mathbf{M}$ and $\mathbf{F}_{\mathbf{L}_1}$, and the latter $N$ groups for separating $\mathbf{F}_{\mathbf{I}_2}$ from $\mathbf{F}^{\rm ini}_{\mathbf{I}_2}$ (obtained by feeding the concatenation of $\mathbf{F}_\mathbf{M}$ and $\mathbf{F}_{\mathbf{I}_1}$ into a $1\times1$ convolutional layer) and $\mathbf{F}_{\mathbf{L}_2}$ (if $\mathbf{L}_2$ is available).
We set $N=4$ and $K=2$ in practice.
Details of the gated language interaction are described as follows.

\noindent\textbf{Adaptive global aggregation module (AGAM).}
As the network structure shown in \Fref{fig:interaction}(a), given an input visual feature $\mathbf{F}^{\rm in}_{\mathbf{V}}\in\mathbb{R}^{h\times w\times C}$ with the spatial resolution of $h\times w$, AGAM is designed for adaptively obtaining a global feature $\mathbf{F}^{\rm glo}_{\mathbf{V}}\in\mathbb{R}^{C}$ that aggregates the contextual information for the subsequent interaction.
The visual feature $\mathbf{F}^{\rm in}_{\mathbf{V}}$ is firstly averaged by an average pooling layer to obtain $\mathbf{F}^{\rm avg}_{\mathbf{V}}\in\mathbb{R}^{C}$.
Then the aggregation process is accomplished via a cross-attention mechanism which contains three linear projection layers with layer normalization~\cite{ba2016layernorm} to conduct query, key, and value projections: $\mathbf{F}^{\rm glo}_{\mathbf{V}} = {\rm Softmax}(\mathbf{Q}\mathbf{K}^\top/\tau)\mathbf{V}$, where the query $\mathbf{Q}\in\mathbb{R}^{C}$ is projected from $\mathbf{F}^{\rm avg}_{\mathbf{V}}$, and the key $\mathbf{K}\in\mathbb{R}^{(hw+1)\times C}$ and value $\mathbf{V}\in\mathbb{R}^{(hw+1)\times C}$ are from the concatenation of $\mathbf{F}^{\rm in}_{\mathbf{V}}$ and $\mathbf{F}^{\rm avg}_{\mathbf{V}}$, and $\tau$ is a learnable scaling factor to control the magnitude of the dot product of $\mathbf{Q}$ and $\mathbf{K}$ before applying the softmax function.


\noindent\textbf{Adaptive global interaction module (AGIM).}
Inspired by existing reflection separation approaches~\cite{wei2019single,hu2023single} which attempt to distinguish transmission and reflection layers in the feature space by feature channel rearrangement (\ie, allocating distinct channels from mixture image features to the two layers), we propose to integrate language interactions at the feature channel level.
To achieve this, as illustrated in~\Fref{fig:interaction}(b), we propose the adaptive global interaction module (AGIM), which employs the channel-wise cross-attention mechanism to conduct interactions between global language features and image features for channel rearrangement.
The query, key, and value projections using linear projection layers with layer normalization~\cite{ba2016layernorm} are conducted on the global language feature $\mathbf{F}_{\mathbf{L}_i}$, the global visual feature $\mathbf{F}^{\rm glo}_{\mathbf{V}}$, and the visual feature $\mathbf{F}^{\rm in}_{\mathbf{V}}$ to generate $\mathbf{Q}\in\mathbb{R}^{C}$, $\mathbf{K}\in\mathbb{R}^{C}$, and $\mathbf{V}\in\mathbb{R}^{C\times hw}$, respectively.
Before the interaction, a language gate is designed to prevent impacts of unavailable language descriptions caused by the recognizable layer ambiguity problem that users may only input one description for the layer recognizable in the mixture image.
Specifically, if the description $\mathbf{L}_i$ corresponding to the current global language feature $\mathbf{F}_{\mathbf{L}_i}$ is available (not set to null), the gate will feed $\mathbf{F}_{\mathbf{L}_i}$ into the following interaction process, otherwise the gate will feed $\mathbf{F}^{\rm glo}_{\mathbf{V}}$, which turns the interaction process to be a channel-wise self-attention.

After filtering by the language gate, an interacted feature can be obtained through the channel-wise cross-attention: $\mathbf{F}^{\rm itr} = {\rm Softmax}(\mathbf{Q}^\top\mathbf{K}/\eta)\mathbf{V}$, where $\eta$ is a learnable scaling factor to control the magnitude of the attention map.
Besides, to adaptively adjust the influence of language guidance based on the correspondence between image and language features, we define another scaling factor $s$ valued as the cosine distance between $\mathbf{Q}$ and $\mathbf{K}$ to multiply with the interacted feature $\mathbf{F}^{\rm itr}$.
Finally, the output of AGIM is obtained by a residual structure: $\mathbf{F}^{\rm out}_{\mathbf{V}} = \mathbf{F}^{\rm in}_{\mathbf{V}} + s\mathbf{F}^{\rm itr}$, which integrates the contextual information from the language description and adjusts features for layer separation.

\subsection{Layer recovery}\label{subsec:recover}

After global language-image interaction and progressive refinement by the gated language interaction stage, we obtain layer features $\mathbf{F}_{\mathbf{I}_i}$ which integrates holistic contextual information from the corresponding language description $\mathbf{L}_i$ (if available).
The layer recovery stage is dedicated to reconstructing each image layer $\mathbf{I}_i\in\mathbb{R}^{H\times W\times 3}$ from its corresponding layer features $\mathbf{F}_{\mathbf{I}_i}$, which is achieved by using individual image decoders.
In the image decoder, the layer feature is firstly upsampled by a transposed convolutional layer with a GELU activation~\cite{hendrycks2016gelu}, then refined by a residual block, and finally projected from the feature domain back into the image domain by a $1\times 1$ convolutional layer with a sigmoid activation, thus accomplishing a precise layer recovery.


\subsection{Loss functions}\label{subsec:loss}

In this section, we introduce the contrastive correspondence loss $\mathcal{L_{\rm ctr}}$ and the layer correspondence loss $\mathcal{L_{\rm lcr}}$ for constraining the proposed method to establish the cross-modality correspondence between language descriptions $\mathbf{L}_i$ and corresponding image layers under the content disturbance from counter layers in mixture images.
We also briefly describe the image layer loss $\mathcal{L_{\rm img}}$ which consists of image- and feature-level loss functions for layer recovery.
We denote the estimated image layers as $\tilde{\mathbf{I}}_i$ and ground truths as $\mathbf{I}_i$.
Details of the loss functions are as follows.

\noindent\textbf{Contrastive correspondence loss.}
CLIP~\cite{radford2021CLIP} conducts contrastive language-image pre-training within batches which successes in establishing cross-modality correspondence between language descriptions and clean images, while our language-guided image reflection separation task requires finding the correspondence between the given language description and a certain layer in the mixture image with the interference of extraneous contents.
To tackle the above issue, we propose a contrastive correspondence loss that conducts contrastive learning between image layers to establish correct cross-modality correspondence.
Specifically, given a language description $\mathbf{L}_i$, the goal of the contrastive correspondence loss is to force the network to learn the relation that the corresponding image layer $\mathbf{I}_i$ is more relevant to $\mathbf{L}_i$ than the counter layer $\mathbf{I}_j$ ($j\neq i$), thus constraining the estimated image layer $\tilde{\mathbf{I}}_i$ to conform to the associated language description $\mathbf{L}_i$.

To measure the relevance between a language description $\mathbf{L}$ and an image layer $\mathbf{I}$, we define a feature-level similarity function $\mathcal{D}(\cdot)$ as:
\begin{eqnarray}
    \mathcal{D}(\mathbf{L},\mathbf{I}) = \sigma(\Psi(\mathbf{F}^{\rm glo}_{\mathbf{L}},\mathbf{F}^{\rm glo}_{\mathbf{I}})),
\end{eqnarray}
where $\sigma$ represents the sigmoid function, $\Psi$ represents the cosine distance, $\mathbf{F}^{\rm glo}_{\mathbf{L}}$ is the global language feature obtained by the language encoder (in \Sref{subsec:extract}), and $\mathbf{F}^{\rm glo}_{\mathbf{I}}$ is the global image feature produced by the AGAM (in \Sref{subsec:interact}).
For an available language description $\mathbf{L}_i$, we calculate its contrastive correspondence loss as: 
\begin{eqnarray}
 \mathcal{L}_{\rm ctr}(\mathbf{L}_i,\tilde{\mathbf{I}}_i,\mathbf{I}_j)=
 -\log(\frac{\mathcal{D}(\mathbf{L}_i,\tilde{\mathbf{I}}_i)}
 {\mathcal{D}(\mathbf{L}_i,\tilde{\mathbf{I}}_i)+\mathcal{D}(\mathbf{L}_i,\mathbf{I}_j)}),
\end{eqnarray}
and we sum up the contrastive correspondence loss of each available language description as the final one. 

\noindent\textbf{Layer correspondence loss.}
We further define a layer correspondence loss to encourage the relevance between the language description $\mathbf{L}_i$ and the estimated image layer $\tilde{\mathbf{I}}_i$ to approach the relevance between $\mathbf{L}_i$ and the ground truth image layer $\mathbf{I}_i$:
\begin{eqnarray}
 \mathcal{L}_{\rm lcr}(\mathbf{L}_i,\tilde{\mathbf{I}}_i,\mathbf{I}_i)={\left\|\mathcal{D}(\mathbf{L}_i,\tilde{\mathbf{I}}_i)-\mathcal{D}(\mathbf{L}_i,\mathbf{I}_i)\right\|}_{1},
\end{eqnarray}
where $\mathcal{D}(\cdot)$ is the same feature-level similarity function as in the contrastive correspondence loss.
We also sum up the layer correspondence loss of each available language description for final supervision.

\noindent\textbf{Image layer loss.}
To achieve high-fidelity recovery of image layers (\ie, transmission and reflection layers), the proposed method is also optimized with loss functions following previous reflection separation methods~\cite{zhang2018single,wan2019corrn,hu2023single}.
Specifically, we utilize loss functions that conduct constraints on the visual quality of estimated images (\ie, the pixel $\mathcal{L_{\rm pix}}$, structural similarity $\mathcal{L_{\rm ssim}}$, and perceptual loss $\mathcal{L_{\rm per}}$) or exploit the inherent relationship between two layers (\ie, the exclusion $\mathcal{L_{\rm exc}}$ and reconstruction loss $\mathcal{L_{\rm rec}}$), and we denote the combination of the above image- or feature-level loss functions as the image layer loss $\mathcal{L_{\rm img}}$\footnote{Details of $\mathcal{L_{\rm img}}$ are provided in the supplementary material.}.

Overall, the total loss function is then formulated as:
\begin{eqnarray}
 \mathcal{L}_{\rm total}=
 \gamma_1\mathcal{L}_{\rm ctr}
 +\gamma_2\mathcal{L}_{\rm lcr}
 +\mathcal{L}_{\rm img},
\end{eqnarray}
where coefficients are set as $\gamma_1=\gamma_2=0.5$.




\subsection{Training strategy}\label{subsec:train}

Due to the recognizable layer ambiguity that sometimes only one layer in the mixture image is recognizable, we propose a randomized training strategy to synergize with the gated language interaction mechanism (in \Sref{subsec:interact}).
Since our training data simulates the recognizable layer ambiguity that some image layers do not have corresponding language descriptions (introduced in the next section), we only feed the available language description corresponding to the other image layer to guide the separation.
For data with descriptions of both layers available, which indicates that both layers are recognizable, we also randomly drop one language description and feed the remaining one into the network to improve the generalization capacity for the proposed method.
In practice, we set the ratio of dropping language descriptions to $30\%$.

We implement the proposed method with PyTorch~\cite{paszke2019pytorch} with a batch size of 16 on two Nvidia GeForce RTX 3090 GPUs.
The model is trained for 40 epochs with Adam optimizer~\cite{kingma2014adam} to update learnable parameters.
Weights are initialized as in~\cite{he2015delving}.
The learning rate is set to $10^{-4}$ initially and decreases to $10^{-5}$ at epoch 30.
\section{Data preparation}

Though existing works have constructed several datasets for single-image reflection separation~\cite{wan2017sir2,zhang2018single,li2020single}, they are unavailable for the proposed language-guided reflection separation framework due to the lack of corresponding language descriptions.
Therefore, we build a dataset containing both synthetic and real data to overcome the data deficiency and facilitate network training and evaluation.
Each group of data is composed of a mixture image, a transmission layer, a reflection layer, and two language descriptions.
Details of synthetic and real data are as follows.

\subsection{Synthetic data}
The synthetic dataset is generated for network training to satisfy the data-driven need of the proposed method.
Due to the demand for paired image-language data, we utilize two prevailing image captioning datasets (\ie, Flickr30k~\cite{young2014flickr30k} and COCO Captions~\cite{chen2015cococaption}) for data generation, which contain 31,000 and 330,000 images respectively, and each image has 5 independent human-generated language descriptions.
We randomly select images from the above two datasets as transmission $\mathbf{T}_{\rm S}$ and reflection scene images $\mathbf{R}_{\rm S}$ and conduct an image synthesis process with linear blending~\cite{hu2023single}:
\vspace*{-0.5\baselineskip}
\begin{equation}
    \hat{\mathbf{M}} = \hat{\mathbf{T}} + \hat{\mathbf{R}} = \alpha\hat{\mathbf{T}}_{\rm S} + \beta\hat{\mathbf{R}}_{\rm S},
\end{equation}
where $\hat{\mathbf{T}}_{\rm S} = g_{\rm inv}(\mathbf{T}_{\rm S})$ and $\hat{\mathbf{R}}_{\rm S} = g_{\rm inv}(\mathbf{R}_{\rm S})$, $g_{\rm inv}$ represents the inverse gamma correction, and $\alpha\in [0.8,1]$ and $\beta\in [0.4, 1]$ are the blending attenuation coefficients as in~\cite{hu2023single}.

Considering the recognizable layer ambiguity that sometimes only one layer is recognizable in a mixture image, we assign a language description only when the corresponding layer is obvious enough in the synthesized mixture image, \ie, $\frac{{\rm mean}(\hat{\mathbf{V}}_l)}{{\rm mean}(\hat{\mathbf{V}}_\mathbf{M})}\geqslant \mu$, where $\hat{\mathbf{V}}_l$ represents the brightness image of $\hat{\mathbf{T}}$ or $\hat{\mathbf{R}}$ in the HSV color space and $\hat{\mathbf{V}}_\mathbf{M}$ represents the brightness image of $\hat{\mathbf{M}}$, and we set $\mu=0.3$ in the data generation process.
Finally, gamma correction is applied to image triplets $\{\hat{\mathbf{T}}, \hat{\mathbf{R}}, \hat{\mathbf{M}}\}$ to obtain $\{\mathbf{T}, \mathbf{R}, \mathbf{M}\}$, and we generate 50000 triplets of data in total for network training.

\subsection{Real data}
Existing real datasets collected for the single-image reflection separation task usually contain mixture images with ground truth of transmission layers (\eg, Zhang~\etal~\cite{zhang2018single} and Nature~\cite{li2020single}), and SIR$^{2}$~\cite{wan2017sir2,wan2022sir2+} further captures ground truths of reflection layers.
For these off-the-shelf real datasets, we augment them by manually adding language descriptions for each group of data to satisfy the input setting of the proposed language-guided reflection separation task.
Specifically, following the annotation principle of COCO Captions~\cite{chen2015cococaption}, we first describe the content of transmission layers in mixture images with entities, attributes (\eg, colors or materials), and relative positions between different entities.
If the content of reflection layers is recognizable in mixture images, we also give language descriptions for reflection layers in the same way.
To further evaluate the generalization capacity of the proposed method, we collect a real dataset (denoted as {\sc RefOL} dataset) containing 100 mixture images from the Internet that are captured in different scenes and with different cameras.
We annotate these mixture images with language descriptions in the same manner as mentioned above.
Following the training strategy of previous methods~\cite{li2020single,dong2021location}, we utilize 200 image pairs from Nature dataset~\cite{li2020single} and 90 pairs from Zhang~\etal~\cite{zhang2018single} for training, and the rest of real data are used for quantitative and qualitative evaluation.
\section{Experiments}
\subsection{Comparison with state-of-the-art methods}

To evaluate the performance of the proposed method, we conduct quantitative and qualitative experiments on existing real datasets~\cite{wan2017sir2,zhang2018single,li2020single} (with our manually annotated language descriptions) and our newly collected dataset {\sc RefOL}.
We compare with state-of-the-art single-image reflection separation methods, including DSRNet~\cite{hu2023single}, YTMT~\cite{hu2021ytmt}, Dong~\etal~\cite{dong2021location}, IBCLN~\cite{li2020single}, CoRRN~\cite{wan2019corrn}, and Zhang \etal~\cite{zhang2018single}.
For fair comparisons, we finetune the above methods on our training data if their training codes are provided.
We report better results between the original pre-trained model and the finetuned version.

\begin{table*}[htbp]
 \caption{Comparison of quantitative results in terms of PSNR~\cite{huynh2008scope} and SSIM~\cite{wang2003multiscale} on real datasets for evaluating the recovery of transmission layers.
 $\uparrow$ ($\downarrow$) indicates larger (smaller) values are better.
 \textbf{Bold} numbers indicate the best-performing results.}
 \vspace{-0.2cm}
  \renewcommand\arraystretch{0.8}
  \centering
  \begin{adjustbox}{width={0.9\textwidth},totalheight={\textheight},keepaspectratio}
  {\begin{tabular}{ccccccccc}
    \toprule
    \multirow{2}[4]{*}{Dataset (size)} & \multirow{2}[4]{*}{Metrics} & \multicolumn{7}{c}{Methods} \\
    \cmidrule{3-9}  & & Zhang~\etal~\cite{zhang2018single} & CoRRN~\cite{wan2019corrn} & IBCLN~\cite{li2020single} & Dong \etal~\cite{dong2021location} & YTMT~\cite{hu2021ytmt} & DSRNet~\cite{hu2023single} & Ours \\
    \midrule
    \multirow{2}[2]{*}{Postcard (199)} & PSNR$\uparrow$ & 20.85 & 22.04 & 23.41 & 23.72 & 22.82 & 24.88 & \textbf{25.02} \\
          & SSIM$\uparrow$ & 0.872 & 0.870 & 0.872 & 0.903 & 0.885 & 0.910 & \textbf{0.915} \\
    \midrule
    \multirow{2}[2]{*}{Object (200)} & PSNR$\uparrow$ & 23.84 & 25.13 & 24.52 & 24.36 & 24.68 & 26.44 & \textbf{26.51} \\
          & SSIM$\uparrow$ & 0.872 & 0.912 & 0.891 & 0.898 & 0.892 & 0.921 & \textbf{0.927} \\
    \midrule
    \multirow{2}[2]{*}{Wild (101)} & PSNR$\uparrow$ & 24.97 & 25.17 & 24.78 & 25.75 & 25.70 & 25.86 & \textbf{26.23} \\
          & SSIM$\uparrow$ & 0.875 & 0.889 & 0.884 & 0.903 & 0.897 & 0.908 & \textbf{0.925} \\
    \midrule
    \multirow{2}[2]{*}{Real20 (20)} & PSNR$\uparrow$ & 22.34 & 21.43 & 21.47 & 23.34 & 23.23 & 23.88 & \textbf{24.05} \\
          & SSIM$\uparrow$ & 0.795 & 0.801 & 0.762 & 0.812 & 0.802 & 0.816 & \textbf{0.824} \\
    \midrule
    \multirow{2}[2]{*}{Nature (20)} & PSNR$\uparrow$ & 20.62 & 20.75 & 23.72 & 23.45 & 21.53 & 22.26 & \textbf{23.87} \\
          & SSIM$\uparrow$ & 0.753 & 0.783 & 0.806 & 0.808 & 0.778 & 0.801 & \textbf{0.812} \\
    \midrule
    \multirow{2}[2]{*}{Average (540)} & PSNR$\uparrow$ & 22.77 & 23.70 & 24.02 & 24.31 & 24.01 & 25.51 & \textbf{25.72} \\
          & SSIM$\uparrow$ & 0.865 & 0.883 & 0.875 & 0.894 & 0.883 & 0.906 & \textbf{0.914} \\
    \bottomrule
    \end{tabular}%
  }
   \end{adjustbox}
 \label{tab:cmp_t}%
\end{table*}

\begin{figure*}[t]
    \centering
    \includegraphics[width=\linewidth]{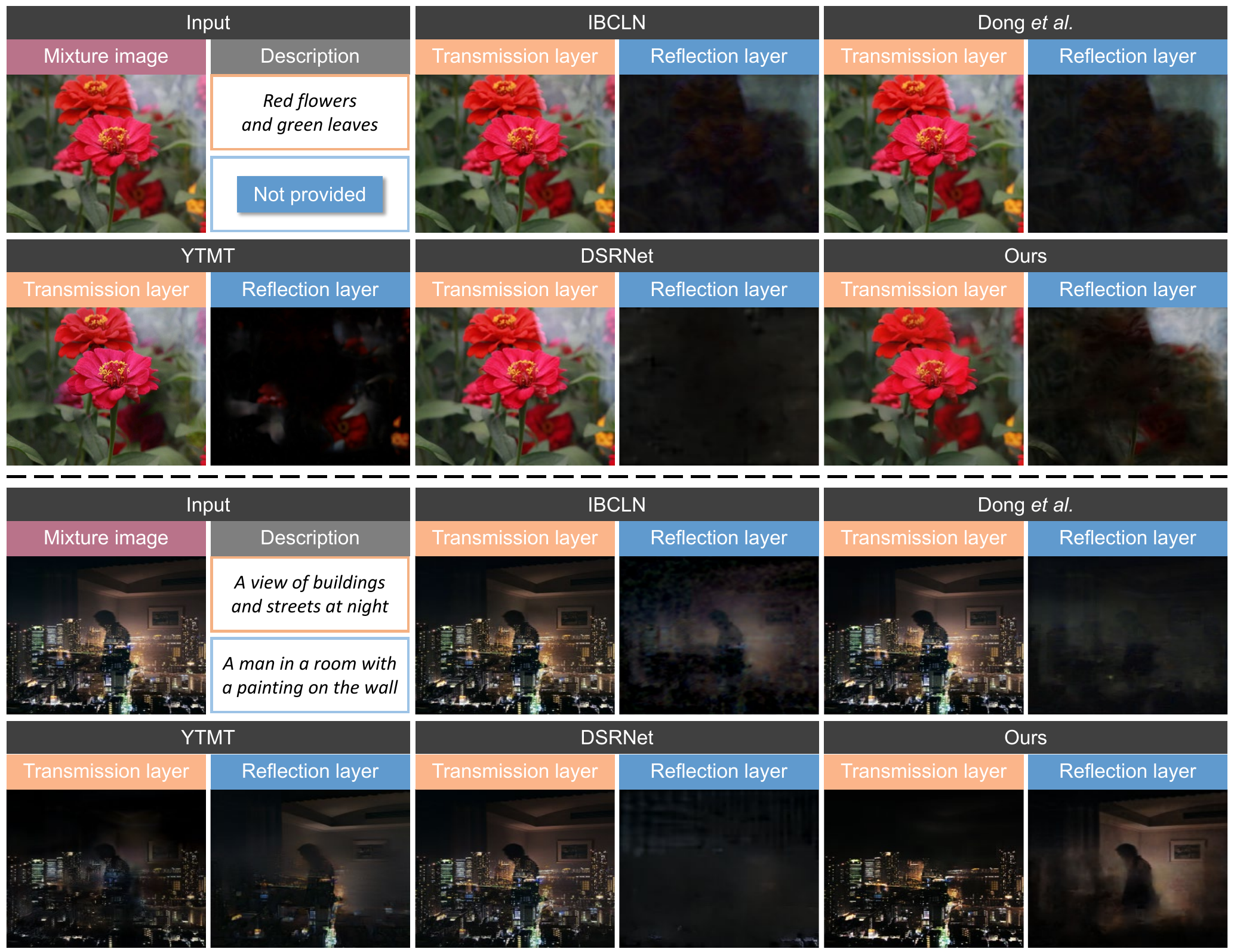}
    \vspace{-0.7cm}
    \caption{Qualitative comparison of estimated transmission and reflection layers on real data, compared with the state-of-the-art methods including DSRNet~\cite{hu2023single}, YTMT~\cite{hu2021ytmt}, Dong~\etal~\cite{dong2021location}, and IBCLN~\cite{li2020single}.
    Please zoom in for details.}
    \label{fig:cmp}
\end{figure*} 

\noindent\textbf{Quantitative comparison.}
Quantitative experiments are conducted on three real datasets for reflection separation, \ie, Nature~\cite{li2020single}, Real20~\cite{zhang2018single}, and three subsets of SIR$^2$~\cite{wan2017sir2} dataset.
Following the setting of existing reflection separation methods~\cite{dong2021location,lyu2019reflection}, we utilize PSNR~\cite{huynh2008scope} and SSIM~\cite{wang2003multiscale} as error metrics for evaluating the recovery of transmission layers\footnote{Quantitative evaluations on the recovery of reflection layers are provided in the supplementary material.}.
As quantitative results shown in~\Tref{tab:cmp_t}, the proposed method achieves the best performance of both PSNR and SSIM, which validates its generalization capacity and the effectiveness of language descriptions.

\noindent\textbf{Qualitative comparison.}
To evaluate the visual quality of reflection separation results, we compare the proposed method with four single-image reflection separation methods, including DSRNet~\cite{hu2023single}, YTMT~\cite{hu2021ytmt}, Dong~\etal~\cite{dong2021location}, and IBCLN~\cite{li2020single}.
Qualitative results on recovering both transmission and reflection layers are shown in \Fref{fig:cmp}. 
As can be observed in \Fref{fig:cmp}, IBCLN~\cite{li2020single} can only remove parts of reflections, while Dong \etal \cite{dong2021location} have trouble in dealing with complex semantic images like the reflections of a man in a room (the second example).
YTMT~\cite{hu2021ytmt} separates layers incorrectly and thus brings the content of transmission layers into reflections.
DSRNet~\cite{hu2023single} fails to recover transmission and reflection layers for these challenging cases.
Contributing to the language guidance, the proposed method generates better visual results and recovers both transmission and reflection layers neatly.


\subsection{Ablation study}

In this section, we conduct several ablation studies with quantitative results shown in \Tref{tab:abl} to investigate the influence of the additional input of language descriptions (denoted as `w/o language'), the network design of AGIM (denoted as `w/o AGIM'), and the loss functions for cross-modality correspondence (denoted as `$\mathcal{L}_{\rm img}$ only').
The performance of the variant `w/o language' suffers from an obvious degradation as we remove language descriptions, which shows the effectiveness of the additional contextual information for layer separation.
The variant `w/o AGIM' replaces AGIMs with simple feature fusion blocks which directly concatenate language and image features and feed them into self-attention blocks, and the decline in performance validates the necessity of our gated interaction mechanism.
The variant `$\mathcal{L}_{\rm img}$ only' is trained with $\mathcal{L}_{\rm img}$, which obtains results slightly better than the variant `w/o language', indicating the significance of establishing the cross-modality correspondence.

\begin{table}[t]\small
  \centering
  \caption{Quantitative results of ablation studies.}
  \vspace{-0.2cm}
  \renewcommand\arraystretch{1.1}
    \begin{tabular}{cccc|c}
    \toprule
    Metrics & w/o language & w/o AGIM & $\mathcal{L}_{\rm img}$ only & Ours \\
    \hline
    PSNR$\uparrow$ & 24.31 & 24.69 & 24.52 & \textbf{25.72} \\
    SSIM$\uparrow$ & 0.885 & 0.901 & 0.893 & \textbf{0.914} \\
    \bottomrule
    \end{tabular}%
  \label{tab:abl}%
  \vspace{-0.2cm}
\end{table}%
\begin{figure}[t]
    \centering
    \includegraphics[width=\linewidth]{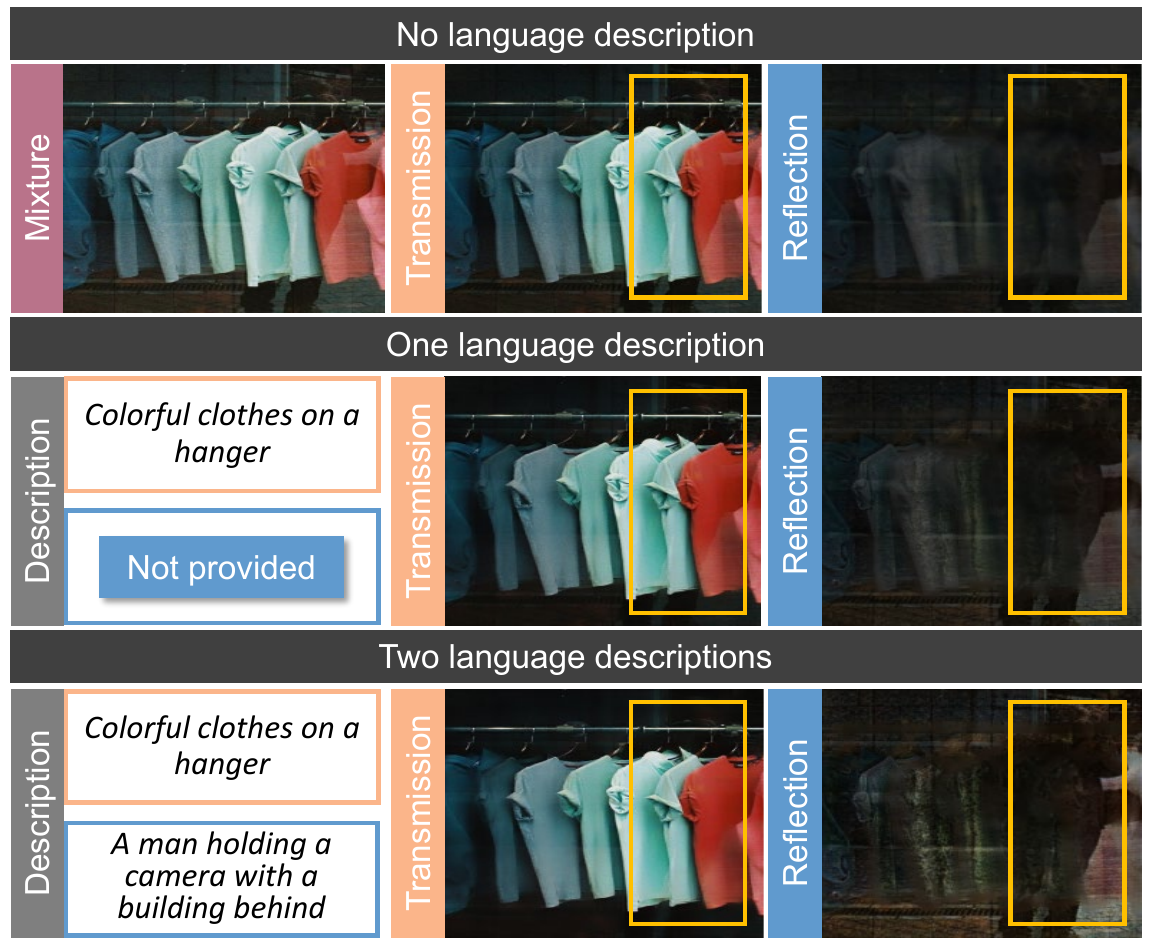}
    \vspace{-0.7cm}
    \caption{Results with different numbers of input language descriptions.}
    \label{fig:abl}
    \vspace{-0.5cm}
\end{figure}

To further verify the effectiveness of the language interaction, we conduct an ablation study by gradually increasing the number of input descriptions.
As shown in yellow boxes of \Fref{fig:abl}, by utilizing more input language descriptions, the separation of transmission (clothes) and reflection layers (a man with a camera) becomes more thorough, which also demonstrates the efficacy and robustness of the proposed method.
\section{Conclusion}

This paper introduces natural language to provide contextual information about image layers for relieving the ill-posed reflection separation problem. 
We develop an end-to-end framework with adaptive global interaction modules and language-image loss functions to effectively manage the modality inconsistency, and we adopt a language gate mechanism with randomized training strategies to handle the recognizable layer ambiguity.
To address data deficiency, a specially built dataset with language annotations significantly aids in training and evaluating the proposed language-guided image reflection separation framework.
Quantitative and qualitative experiments on real data demonstrate the effectiveness of introducing language descriptions for reflection separation.

\noindent\textbf{Limitations}.
The proposed method may not distinguish transmission and reflection layers accurately when their contents are similar.
For such ambiguous cases, a more flexible language-guided mechanism is needed.
This might be solved by exploring a better interaction approach, which is left as our future work.

\section*{Acknowledgement}
This work is supported by National Natural Science Foundation of China under Grant No. 62136001, 62088102.

{
    \small
    \bibliographystyle{ieeenat_fullname}
    \bibliography{main}
}


\end{document}